\documentclass[12pt]{article}
\usepackage[top=1in, bottom=1in, left=1in, right=1in]{geometry}
\usepackage{natbib}
\usepackage{amsmath}
\usepackage{amssymb}
\usepackage{latexsym}
\usepackage{sectsty}
\usepackage{amsfonts}
\usepackage{epsfig}
\usepackage{subfigure}
\usepackage{setspace}
\usepackage{threeparttable}
\usepackage{booktabs}
\usepackage{url}
\usepackage{microtype}
\usepackage{fixmath}
\usepackage{graphicx}
\usepackage{algorithm}
\usepackage{algorithmic}
\allsectionsfont{\sffamily\mdseries}
\paragraphfont{\sffamily\bfseries}

\newcommand{\blambda}{\boldsymbol{\lambda}}

\graphicspath{{./figures/}}

\title{Stochastic Annealing for Variational Inference}
\author{San Gultekin, Aonan Zhang and John Paisley\\Department of Electrical Engineering\\ Columbia University}
\date{}

\begin{document}

\maketitle

\begin{abstract}
We empirically evaluate a stochastic annealing strategy for Bayesian posterior optimization with variational inference. Variational inference is a deterministic approach to approximate posterior inference in Bayesian models in which a typically non-convex objective function is locally optimized over the parameters of the approximating distribution.  We investigate an annealing method for optimizing this objective with the aim of finding a better local optimal solution and compare with deterministic annealing methods and no annealing. We show that stochastic annealing can provide clear improvement on the GMM and HMM, while performance on LDA tends to favor deterministic annealing methods.
\end{abstract}

\onehalfspacing

\section{Introduction}
Machine learning has produced a wide variety of useful tools for addressing a number of practical problems, often for those which involve large-scale datasets. Indeed, a number of disciplines ranging from recommender systems to bioinformatics rely on machine intelligence to extract useful information from their datasets in an efficient manner. One of the core machine learning approaches to such tasks is to define a prior over a model on data and infer the model parameters through posterior inference \citep{Blei:2014}. The gold-standard in this direction is Markov chain Monte Carlo (MCMC), which gives a means for collecting samples from this posterior distribution in an asymptotically correct way \citep{Robert:2004}. 

A frequent criticism of MCMC is that it is not scalable to large data sets---though recent work has begun to address this (e.g., \cite{Welling:2011,Maclaurin:2014}). Instead, variational methods \citep{Wainwright:2008} are proposed as an alternative for approximating the posterior distribution of a model more quickly by turning inference into an optimization problem over an objective function. Though the learned distribution is not as technically correct as the empirical distribution constructed from MCMC samples, fewer iterations are required and ideas from stochastic optimization are immediately applicable for large-scale inference \citep{Hoffman:2012}.

However, a significant issue faced by variational inference methods is that the objective is usually highly non-convex, and so only locally optimal solutions of the posterior approximation can be found. One  response to this problem is to simply rerun the optimization from various random starting points and select the best local optimal solution. This opens variational inference up to the same criticisms as MCMC, since the cumulative number of iterations performed by variational inference may be comparable to a single chain of MCMC. Therefore, the advantage of scalability with variational inference is significantly reduced.

Since variational inference is an instance of non-convex optimization, trying to improve this local optimal problem with the existing annealing approaches is a promising direction. Deterministic annealing has been studied for variational inference both formally \cite{Katahira:2008,Yoshida:2010,Abrol:2014} and informally \cite{Beal:2003}. These approaches perform a deterministic inflating of the variational entropy term, which shrinks with each iteration to allow for exploration of the variational objective in early iterations. Quantum annealing has been studied as well \cite{Sato:2009}.

Another long-studied annealing approach involves stochastic processes and have been introduced and analyzed in the context of global minimization of a non-convex function \citep{Benzi:1982,Kirkpatrick:1983,Cerny:1985,Geman:1986}. Though the conditions for finding such a global optimum may be impractical, the resulting theoretical insights have suggested practical methods for finding better local optimal solutions than found by their non-annealed counterparts. Unlike deterministic annealing, stochastic annealing appears to have been overlooked for variational inference. With this motivation, the goal of this paper is to develop a stochastic annealing algorithm for variational inference and compare its performance with deterministic and non-annealed optimization.

We demonstrate that, like deterministic annealing, improving the performance of variational inference without compromising its scalability is possible using stochastic annealing. Our approach is inspired by the method of simulated annealing \citep{Kirkpatrick:1983}, which prevents the gradient steps from getting trapped in a bad local optimum early on. We show that this approach can improve the performance of variational inference on several models, often improving over deterministic annealing.

The rest of the paper is organized as follows. In Section 2, we present an overview of annealing for optimization and how it can be connected to variational inference. In Section 3 we present our method in the context of conjugate exponential family models. In Section 4 we validate our approach with three models: Latent Dirichlet allocation \citep{Blei:2003b}, the hidden Markov model \citep{Rabiner:1989} and the Gaussian mixture model \citep{Bishop:2006}.

\section{Background}
\subsection{Variational inference}
Given data $X$ and a model with variables $\Theta = \{\theta_i\}$, the goal of posterior inference is to find $p(\Theta|X)$. This is intractable in most models and so approximate methods are used. Mean-field variational inference \citep{Wainwright:2008} performs this task by proposing a simpler factorized distribution $q(\theta) = \prod_i q(\theta_i)$ to approximate $p(\Theta|X)$ by minimizing their KL-divergence. This is equivalently done by maximizing the objective function
\begin{equation}
 \mathcal{L} = \mathbb{E}_q[\ln p(X,\Theta)] - \mathbb{E}_q[\ln q].
\end{equation}
Computing the objective only requires the joint likelihood $p(X,\Theta)$, which is known by definition and is a function of the parameters of each $q(\theta_i)$, $\blambda = \{\lambda_i\}$.

The function $\mathcal{L}$ can be optimized using gradient ascent on the parameters of $q$, which for step $t$ can be written as
\begin{equation} \label{vargrd}
\blambda_{t+1} \gets \blambda_{t} + \gamma_t \nabla \mathcal{L}|_{\blambda_t}.
\end{equation}
where $\gamma_t$ is a step size. In practice this gradient is usually done for each $\lambda_i$ separately holding the others fixed rather than for the entire set $\blambda$.

Solving the approximate inference problem with variational methods has proved useful in a number of applications, but one important shortcoming is that the updates in \eqref{vargrd} are only guaranteed to converge to a local maximum for non-convex problems. Clearly it is important to come up with optimization procedures that find better local optima, or even the global optimum solution. The updates in \eqref{vargrd} arise in a many problems where optimization by gradient methods is used; in these areas, research has been done toward finding better local optima that can be modified for application to variational inference as well \citep{Benzi:1982,Kirkpatrick:1983,Cerny:1985,Geman:1986}.

\subsection{Simulated annealing}
Since \eqref{vargrd} always steps in the direction of the gradient the result will get trapped in a local optima that is highly dependent on the initialization. One way to overcome this problem is the method of simulated annealing \citep{Kirkpatrick:1983}. The basic idea here is to instead make the update
\begin{equation} \label{ann}
\blambda_{t+1} \gets \blambda_{t} + \gamma_t \nabla \mathcal{L}|_{\blambda_t} + T_t \varepsilon_t~,
\end{equation}
where $\varepsilon_t$ is a random noise vector controlled by a ``temperature'' variable $T_t \geq 0$ that converges to zero as $t \to \infty$. The update is then accepted or rejected in a manner similar to Metropolis-Hasting MCMC. The idea is that, in the initial steps the value of $t$ is large enough to prevent $\blambda_t$ from getting trapped in a local maximum (i.e., $\blambda_t$ is volatile enough to escape from the local maximum due to the high temperature $T_t$). As the temperature decreases the movement is more restricted to being ``uphill'' until the sequence eventually converges.

Simulated annealing was first used for discrete variables \citep{Kirkpatrick:1983,Cerny:1985,GemanGeman}. This was later extended to continuous random variables and analyzed in the context of continuous-time processes \citep{Geman:1986}, which results in the following Langevin-type Markov diffusion,
\begin{equation} \label{ctcs}
d\blambda(t) = \nabla \mathcal{L} dt + T(t) d\varepsilon(t)~,
\end{equation}
where $\varepsilon(t)$ is a standard multi-dimensional Brownian motion. \citet{Geman:1986} and \citet{Chiang:1987} showed how, under certain conditions, this process concentrates at the global maximum of $\mathcal{L}$ as $T \to 0$. \citet{Kushner:1987} and \citet{Gelfand:1991,Gelfand:1993} later developed discrete-time versions of this that have the same convergence property. Ideas related to simulated annealing have proved useful in machine learning research from the perspective of MCMC sampling. For example, in Hamiltonian Monte Carlo \citep{Neal:HMC} and sampling with gradient Langevin dynamics \citep{Welling:2011,Welling:2012} gradient information is combined with noise to produce more efficient sampling.

These results suggest that simulated annealing can significantly improve the performance of gradient-based optimization. With that said, they are also limited in the sense that: ($i$) The injected noise is restricted to have a Gaussian distribution, and ($ii$) choosing the optimal cooling function $T(t)$ is often impractical. In addition, in the variational setting evaluating the objective function to accept/reject may be a time-consuming procedure. To this end, modified annealing procedures that are outside the realm of provable convergence may still be useful for practical problems \citep{Geman:1986}, and one may trade guaranteed convergence with practicality. In this case, the global optimum is traded for a better local optimum than those found by non-annealed gradient ascent.

\section{Annealing for Variational Inference}
 We describe our ``practical'' modification to the globally convergent simulated annealing algorithm in the context of variational inference for conjugate exponential models.

\subsection{Variational inference for CEF models}
Variational inference for conjugate exponential family models, in which $q$ is in the same family as the prior and $\lambda_i$ is the natural parameter for $q(\theta_i)$, allows the gradient $\nabla_{\lambda_i}\mathcal{L}$ to be written in a simple form,
\begin{equation}\label{eqn.gradL}
 \nabla_{\lambda_i}\mathcal{L} = -\left(\frac{d^2\ln q(\theta_i)}{d\lambda_i d\lambda_i^T}\right)(\mathbb{E}_q[\boldsymbol{t}] + \lambda_0 - \lambda_i).
\end{equation}
The vector $E_q[\boldsymbol{t}]$ is the expected sufficient statistics of the conditional posterior $p(\theta_i|X,\Theta_{-i})$ using all other $q$ distributions and $\lambda_0$ is from the prior on $\theta_i$. 

Using a positive definite matrix $\boldsymbol{M}$, the gradient update $\lambda_i \gets \lambda_i + \gamma_t \boldsymbol{M}\nabla_{\lambda_i}\mathcal{L}|_{\lambda_i}$ is globally optimal for a particular $\lambda_i$ conditioned on all other $q$ distributions when $\gamma_t = 1$ and $\boldsymbol{M}= -\left(d^2\ln q(\theta_i)/d\lambda_i d\lambda_i^T\right)^{-1}$. This corresponds to setting the gradient in Eq.\ (\ref{eqn.gradL}) to zero which gives the familiar update 
\begin{equation}
 \lambda_i \gets \mathbb{E}_q[\boldsymbol{t}] + \lambda_0.
\end{equation}
To develop stochastic annealing, our is to modify this update in a manner similar to the transition from Eq.\ (\ref{vargrd}) to Eq.\ (\ref{ann}). 

Variational inference also requires initializing the variational parameters of each $q(\theta_i)$ distribution. In this paper, we assume that each $\theta_i$ is initialized randomly in an appropriate way.

\subsection{Deterministic annealing for VI}
Deterministic annealing has been proposed for variational inference \cite{Katahira:2008}. This gives a general framework for annealing the variational objective function that does not involve any randomness. With deterministic annealing, a trade off is made between the entropy and the expected log joint likelihood to avoid being trapped in a bad local optimum early on. This is done by multiplying the entropy term in the variational lower bound by a ``temperature'' parameter $T>1$,

$$\mathcal{L} = \mathbb{E}_q [\text{ln}~p(X,\Theta)] - T_t \mathbb{E}_q[\text{ln}~q].$$

In early iterations (indexed by $t$) larger values of $T$ favor smoother distributions because such distributions have higher entropy, and thus a higher-value for the objective function. As the number of iterations increase, $T$ is gradually lowered (or ``cooled'') which lets the variational distribution fit to the data. This way, better values for the variational parameters can be obtained. 

We can take the derivative of the lower bound with respect to $\lambda_i$ to find the optimal update. This gives
\begin{equation}\label{eqn.gradL}
 \nabla_{\lambda_i}\mathcal{L} = -\left(\frac{d^2\ln q(\theta_i)}{d\lambda_i d\lambda_i^T}\right)(\mathbb{E}_q[\boldsymbol{t}] + \lambda_0 - T_t\lambda_i).
\end{equation}
Pre-multiplying by $\boldsymbol{M}= -\left(d^2\ln q(\theta_i)/d\lambda_i d\lambda_i^T\right)^{-1}$ as before gives
\begin{equation}
 \lambda_i \gets \frac{1}{T_t}(\mathbb{E}_q[\boldsymbol{t}] + \lambda_0).
\end{equation}
As is evident, deterministic annealing down-weights the amount of information in the posterior, thus increasing the entropy, but the information it does incorporate is determined by the data. 

\subsection{Stochastic annealing for VI}
Motivated by Eq.\ (\ref{ann}), we propose a different approach to annealing the variational objective function. Similar to that equation, we propose the annealing update
\begin{equation}
 \lambda_i \gets \lambda_i + \gamma_t \boldsymbol{M}\nabla_{\lambda_i}\mathcal{L}|_{\lambda_i} + T_t\varepsilon_t.
\end{equation}
We chose the form of the preconditioning matrix $\boldsymbol{M}$ and the noise $\varepsilon_t$ out of convenience, and also re-parameterize $T_t$ as follows,
\begin{equation}
 \boldsymbol{M}= -\left(\frac{d^2\ln q(\theta_i)}{d\lambda_i d\lambda_i^T}\right)^{-1},
\end{equation}
\begin{equation}
 T_t =  \gamma_t\rho_t,\quad\quad \varepsilon_t = \eta_t - \mathbb{E}_q[\boldsymbol{t}] - \lambda_0.
\end{equation}
We set $\rho_t$ to be a step size that is shrinking to zero as $t$ increases and discuss the random vector $\eta_t$ shortly. Using the optimal setting of $\gamma_t$ for conjugate exponential models discussed above, we set $\gamma_t = 1$ for all $t$, which gives the convenient update
\begin{equation}\label{eqn.savi_up}
 \lambda_i \gets (1-\rho_t)(\mathbb{E}_q[\boldsymbol{t}] + \lambda_0) + \rho_t \eta_t.
\end{equation}
In contrast to simulated annealing, and similar to \citet{Welling:2011}, we assume that all updates are accepted with probability one to significantly accelerate inference. The step size $\rho_t$ is a value decreasing to zero, and in this paper we assume that $\rho_t = 0$ for all $t > T$, with $T$ preset. Therefore, this assumption does not impact convergence of the algorithm to a local optimal solution. We evaluate the quality assuming probability one acceptance by our experiments. 

We see that there is some relationship between stochastic and deterministic annealing. In deterministic annealing, $T_t > 1$ and decreasing to one. The value $T_t = (1-\rho_t)^{-1}$ is one possible setting, and so the first term in Eq. (\ref{eqn.savi_up}) can be viewed as exactly deterministic annealing. In addition, we introduce a random term, which has the effect of again reducing the entropy of $q$, but to a perturbed location that allows for exploration of the objective function similar to deterministic annealing.

We observe that this annealing method requires setting $\eta_t$ at each iteration. Recalling that $\lambda_i$ is randomly initialized according to an appropriate method, we propose generating $\eta_t$ according to the same random initialization. In this case, each update has the intuitive interpretation of being a weighted combination of the true model updates and a brand new random initialization. As $t$ increases, the weight of the initialization decreases to zero until the correct updates are used exclusively. We present an outline of this simulated annealing method in Algorithm \ref{alg.aVI}.

\begin{algorithm}[t]
\caption{An annealing algorithm for VI}\label{alg.aVI}
\begin{algorithmic}[1]
\STATE For conjugate exponential models with $q$ distributions in the same family as the prior.
\STATE Randomly initialize natural parameters $\lambda_i$ of $q(\theta_i)$.\vspace{1pt}
\FOR{each $q(\theta_i)$ in iteration $t$}\vspace{1pt}
\STATE Set the step size $\rho_t$.
\STATE Calculate expected sufficient statistics $\mathbb{E}_q[\boldsymbol{t}]$.
\STATE Generate new random initialization $\eta_{i,t}$ for $\lambda_i$.
\STATE Update $\lambda_i \gets (1-\rho_t)(\mathbb{E}_q[\boldsymbol{t}] + \lambda_0) + \rho_t \eta_{i,t}$.
\ENDFOR
\end{algorithmic}
\end{algorithm}

\section{Experiments}
We evaluate our annealing approach for variational inference using three models: Latent Dirichlet allocation, the discrete hidden Markov model and the Gaussian mixture model. We compare the performance of stochastic annealing (stochAVI) with deterministic annealing (detAVI) and no annealing (VI). For deterministic annealing, we follow the approach of \citet{Katahira:2008}, with the specific extension of this to LDA discussed in \citet{Abrol:2014}.  

We describe the setup, annealing strategy and results for each of these models below. In each section, we first briefly review the problem setup, including the model variables and selected $q$ distributions. We then discuss how our annealing approach can be applied to the problem. Finally, we discuss the results on the model. For all experiments we set $\rho_t = 0.9^t$ for stochastic annealing and $T_t = 5(1-\rho_t)^{-1}$, which we empirically found to given results representative of the two methods; we note that the performance did not change significantly around the numbers $0.9$ and $5$. We mention that, due to the minimal overhead, the running time for stochAVI and detAVI was essentially the same as for VI.

\subsection{Latent Dirichlet allocation}

\paragraph{Setup.} We first present experiments on a text modeling problem using latent Dirichlet allocation (LDA). We consider the four corpora indicated in Table \ref{tab.docs}. The model variables for a $K$-topic LDA model of $D$ documents are $\Theta = \{\beta_{1:K},\pi_{1:D}\}$. The vector $\pi_d$ gives a distribution on $\boldsymbol{\beta}$ for document $d$ and each topic $\beta_k$ is a distribution on $V$ vocabulary words. We use the factorized $q$ distribution $q(\beta_{1:K},\pi_{1:D}) = \Big[\prod_k q(\beta_k)\Big]\Big[\prod_d q(\pi_d)\Big]$, and set each to be Dirichlet, which is the same family as the prior. We set the Dirichlet prior parameter of $\pi_d$ to $1/K$ and the Dirichlet prior parameter of $\beta_k$ to $100/V$. We initialize all $q$ distributions by scaling up a uniform Dirichlet random vector, with the specific scaling discussed below.

\paragraph{Annealing.}
The standard variational parameter updates for LDA involve summing expected counts over all words and documents. This is done by introducing an additional variational distribution $\phi_{d,n}$ on the allocation probability of the topic associated with the $n$th word in the $d$th document. Below, we focus on the update of $q(\beta_k)$, and noting that a simple modification is required for $q(\pi_d)$. We recall that the update for the variational parameter $\lambda_k$ of $\beta_k$ is
\begin{equation}
 \lambda_k \leftarrow \textstyle \sum_{d,n} \phi_{d,n}(k)w_{d,n} + \lambda_0,
\end{equation}
where $w_{d,n}$ is an indicator vector of length $V$ for word $n$ in document $d$. Using an un-scaled initialization of $\eta_{k,t}/\mbox{scale} \sim \mbox{Dir}(1,\dots,1)$, we modify this update to
\begin{equation}
 \lambda_k \leftarrow \textstyle (1-\rho_t)(\sum_{d,n} \phi_{d,n}(k)w_{d,n} + \lambda_0) + \rho_t \eta_{k,t}.
\end{equation}
As discussed above, with this update we first form the correct update to $\lambda_k$ using the data. We then generate a new initialization for $\lambda_k$ and take a weighted average of the two vectors using the step size $\rho_t$. We set $\mbox{scale} = cD/K$ for updating $q(\beta_k)$ and $\mbox{scale} = c/K$ for updating $q(\pi_d)$, where $c$ is the average number of words per document.

\begin{table}
\caption{The four corpora used in the LDA experiments and their relevant statistics.}\vspace{10pt}\label{tab.docs}\centering
\begin{tabular}{ |l|cccc| }
\hline
& NIPS & ArXiv & NYT & HuffPost \\
 \hline
 \# docs & 2.5K & 3.8K & 8.4K & 4K\\ 
 \# vocab  & 2.2K & 5K & 3.0K & 6.3K\\
 \# tokens & 2.5M & 234K & 1.2M & 906K\\
 \# word/doc & 1000 & 62 & 143 & 226\\
\hline
\end{tabular}
\end{table}

\begin{figure}[t!]\centering
\subfigure[Huffington Post]{\includegraphics[width=.45\textwidth]{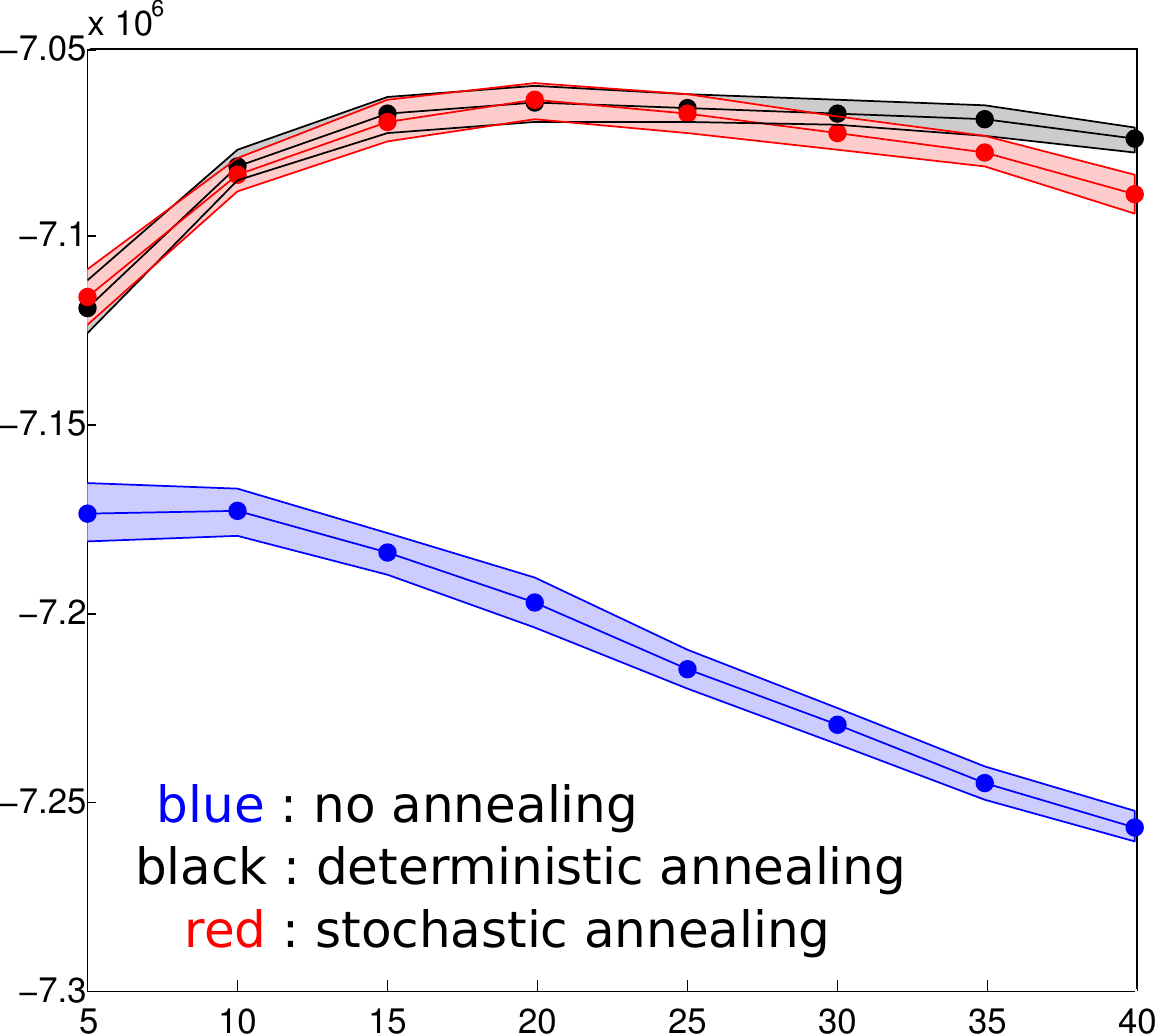}}
\subfigure[arXiv]{\includegraphics[width=.45\textwidth]{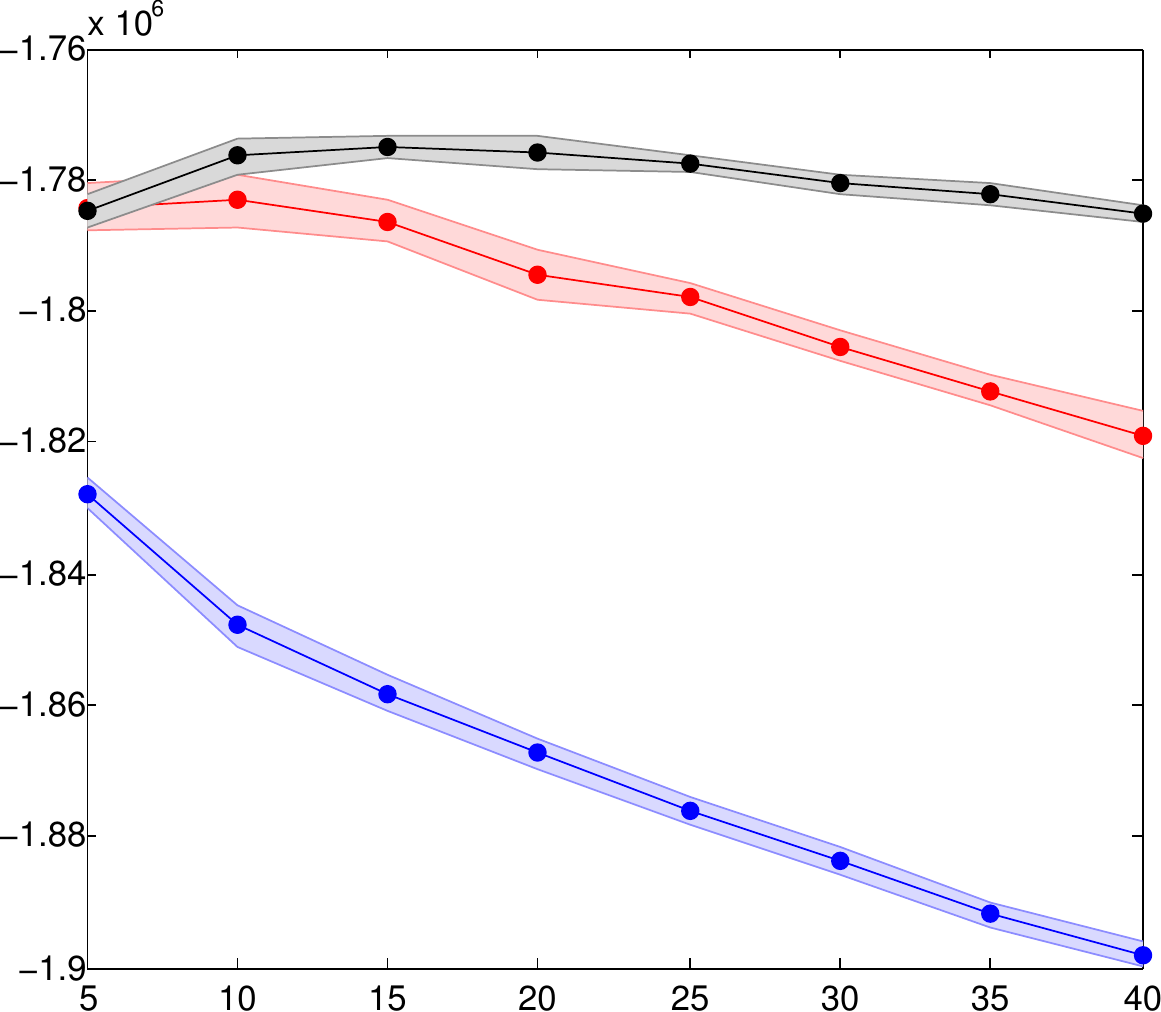}}
\subfigure[New York Times]{\includegraphics[width=.45\textwidth]{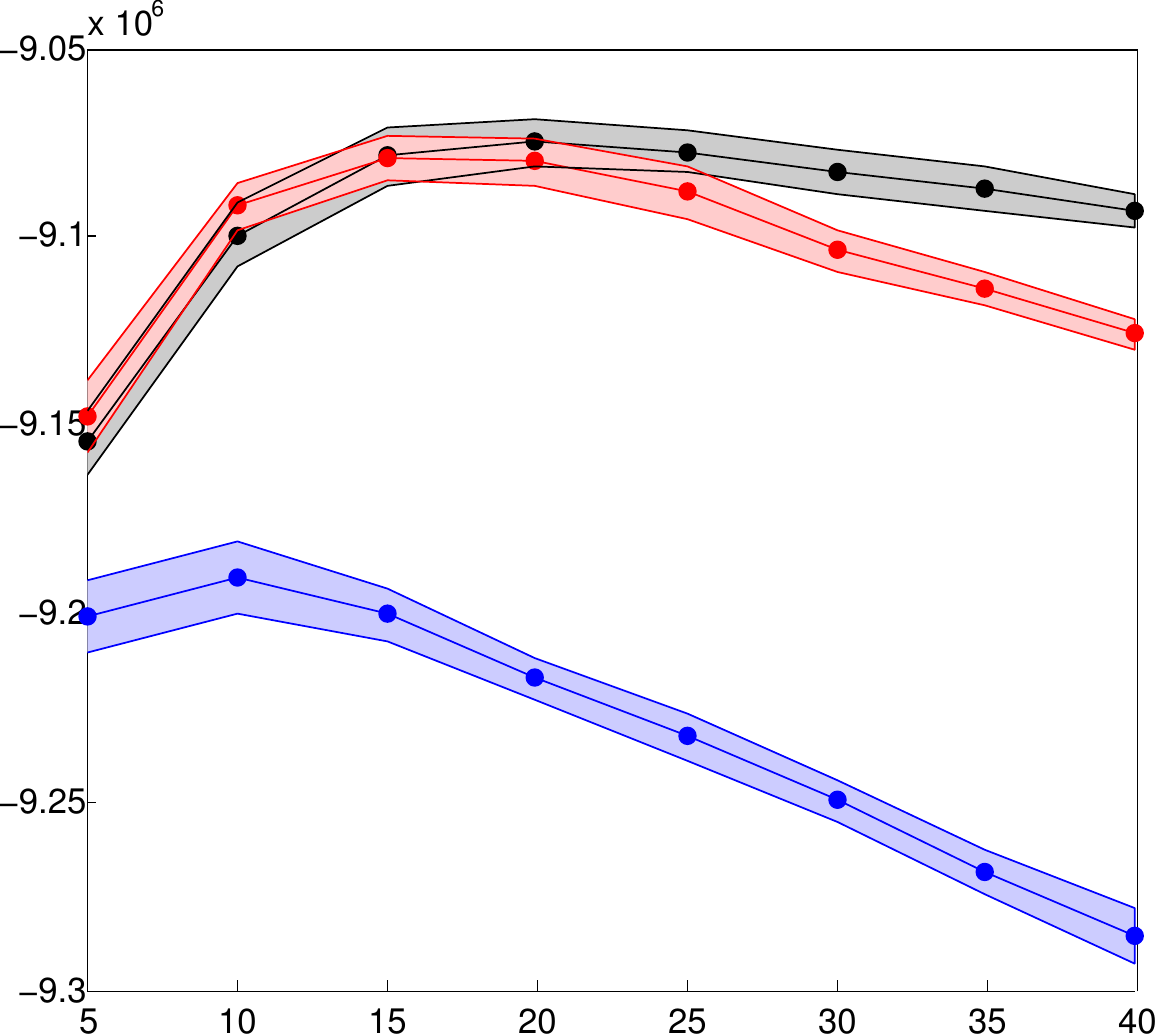}}
\subfigure[NIPS]{\includegraphics[width=.45\textwidth]{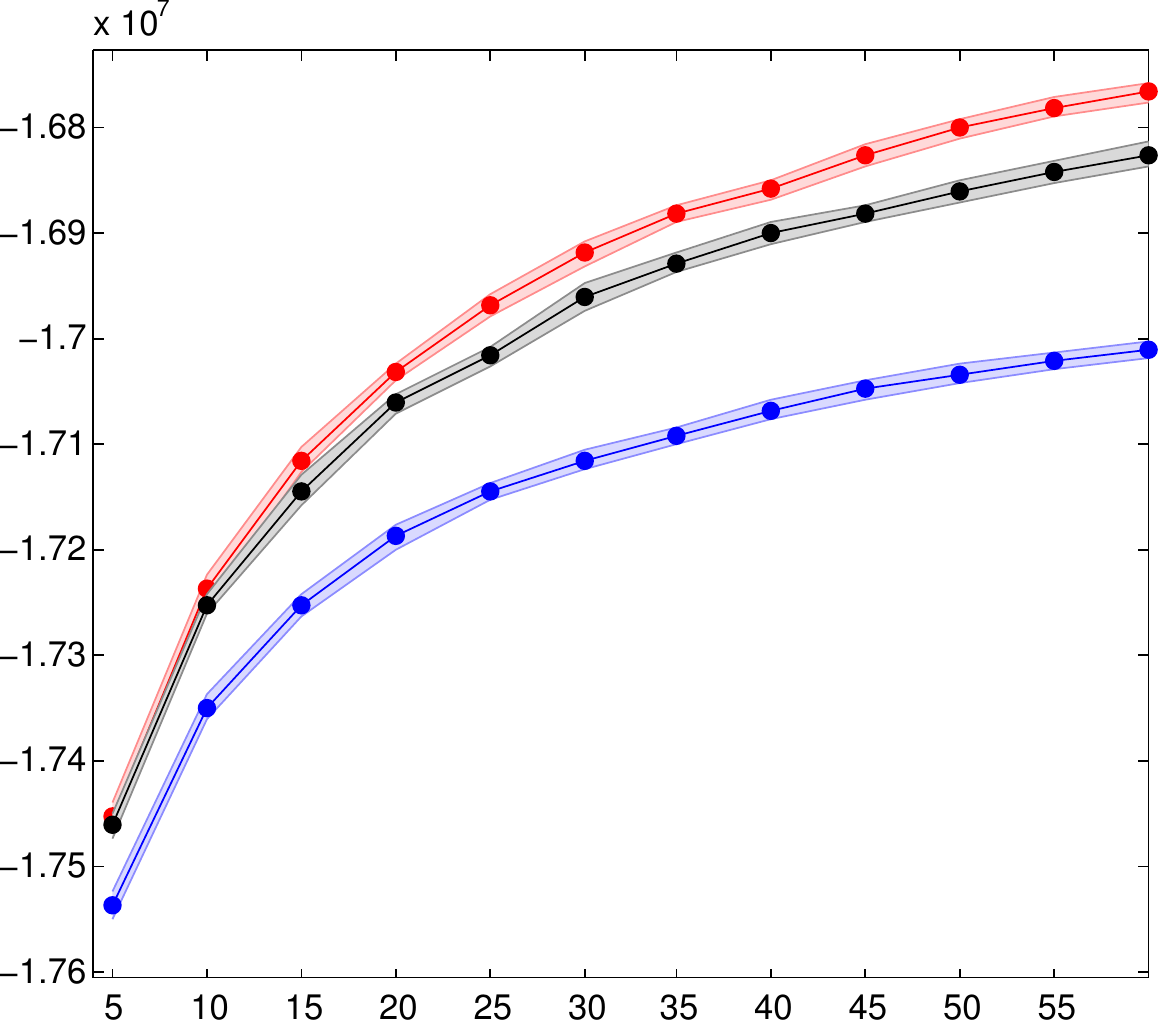}}
\caption{The variational objective function vs number of topics for variational inference using stochastic, deterministic and no annealing for LDA. In general, deterministic annealing outperforms our method, with some exceptions. Both annealing methods significantly outperform no annealing. We observe that annealing provides different, and possibly more accurate information on the appropriate number of topics when using the lower bound for model selection.}\label{fig.boundLDA}
\end{figure}

\paragraph{Results.} In Figure \ref{fig.boundLDA} we show plots of the final value of the variational objective as a function of $K$ using 50 runs of each model and inference method. As is evident, stochAVI is not uniformly better than detAVI in this problem, though both are clearly superior to VI without annealing. Empirically, we see that the performance of stochAVI improves compared with detAVI as the number of words per document increases, which is consistent with our additional experiments not shown here. This indicates a regime in which our approach would be preferable. We also observe that the annealed and non-annealed methods disagree on the appropriate number of topics for each corpus. Since the lower bound can be used for performing this model selection, these results indicate that the lower bound of the marginal likelihood given by the variational objective does not necessarily peak at the same value of $K$ as the true marginal likelihood. Since the annealed results are overall better, they can be 
considered as 
providing better justification for choosing $K$.

\subsection{The discrete hidden Markov model}

\paragraph{Setup.} For the next experiment we considered the discrete $K$-state hidden Markov model (HMM). The model variables are $\Theta = \{\pi,A,B\}$, where $\pi$ is an initial state distribution, $A$ is the Markov transition matrix and the rows of matrix $B$ correspond to the emission probability distributions for each state. All priors are Dirichlet distributions and we therefore use Dirichlet $q$ distributions for the factorization $q(\pi,A,B) = q(\pi)\prod_{k=1}^K q(A_{k,:})q(B_{k,:})$. For the priors on $A$ and $\pi$ we set the Dirichlet parameter to $1/K$. For the priors on $B$ we set the Dirichlet parameter to $10/V$, where $V$ is codebook size. As with LDA, we initialize all $q$ distributions by scaling up a uniform Dirichlet random vector to the data size.  

We evaluate the annealed and standard versions of variational inference on two datasets: A character trajectories dataset from the UCI Machine Learning Repository, and the Million Song Dataset (MSD). The characters dataset consists of sequences of spatial locations of a pen as it is used to write one of 20 different characters. There are 2,858 sequences in total from which we held out 200 for testing (ten for each character). We quantized the 3-dimensional sequences using a codebook of size 500 learned with K-means. For MSD we quantized MFCC features using 1024 codes and extracted sequences of length 50 from 500 different songs.

\begin{figure}[h!]
\centering
\subfigure[5-state hidden Markov model]{\includegraphics[width=.77\textwidth]{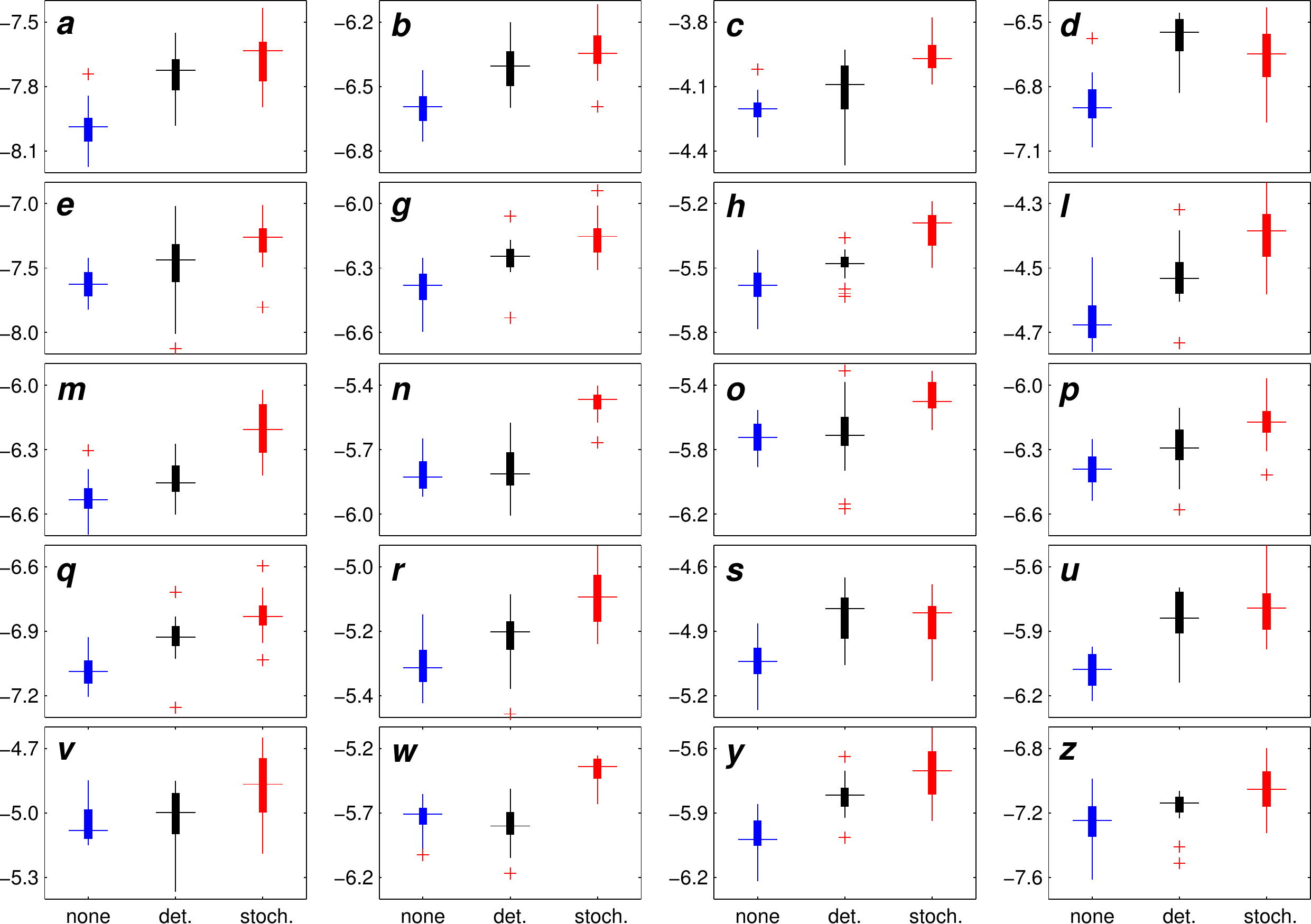}}
\subfigure[10-state hidden Markov model]{\includegraphics[width=.77\textwidth]{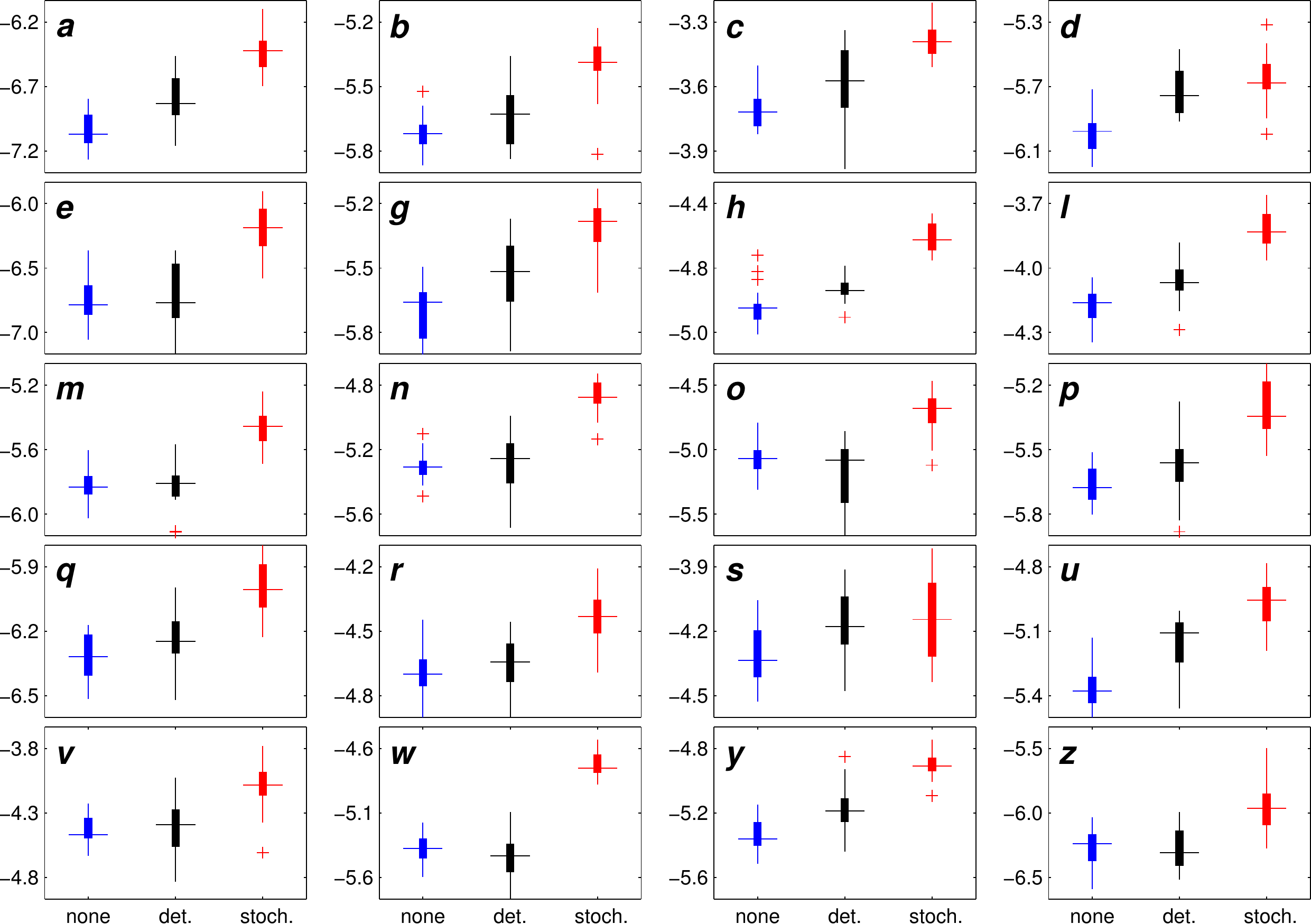}}
\caption{The variational objective function ($\times 10^4$) for each character. (red) stochAVI, (black) detAVI, (blue) VI. The proposed annealing consistently converges to a better local optimal approximation to the posterior of the hidden Markov model.}\label{fig.digit_bound}
\end{figure}

\paragraph{Annealing.} As with LDA, the update for each $q$ involves a sum over expected counts, this time involving the state transition probabilities learned from the forward-backward algorithm. Very generally speaking these updates are of the form
\begin{equation}
 \lambda_k \leftarrow \textstyle \sum_n\sum_m \phi_{nm,k} + \lambda_0,
\end{equation}
where $\lambda_0$ is a prior and $\phi_{nm,k}$ is a probability relating to the $m$th emission in sequence $n$ and state $k$, which is calculated by introducing a variational multinomial $q$ distribution on the hidden data of state transitions. Since the distributions used are the same, the annealed modification is essentially identical to LDA. Using an un-scaled initialization of $\eta_k/\mbox{scale} \sim \mbox{Dir}(1,\dots,1)$, we modify this update to
\begin{equation}
 \lambda_k \leftarrow \textstyle (1-\rho_t)(\sum_n\sum_m \phi_{nm,k} + \lambda_0) + \rho_t \eta_{k,t}.
\end{equation}
That is, we form the correct update to $\lambda_k$ using the data, generate a new initialization for $\lambda_k$ and then take a weighted average of the two using a step size $\rho_t \rightarrow 0$. We again set $\rho_t = 0.25\max(0,1-t/50)$ and set $\mbox{scale} = cN/K$, where $N$ is the number of sequences and $c$ is the expected length of a sequence.

\begin{figure}
\centering
 \includegraphics[width=.6\textwidth]{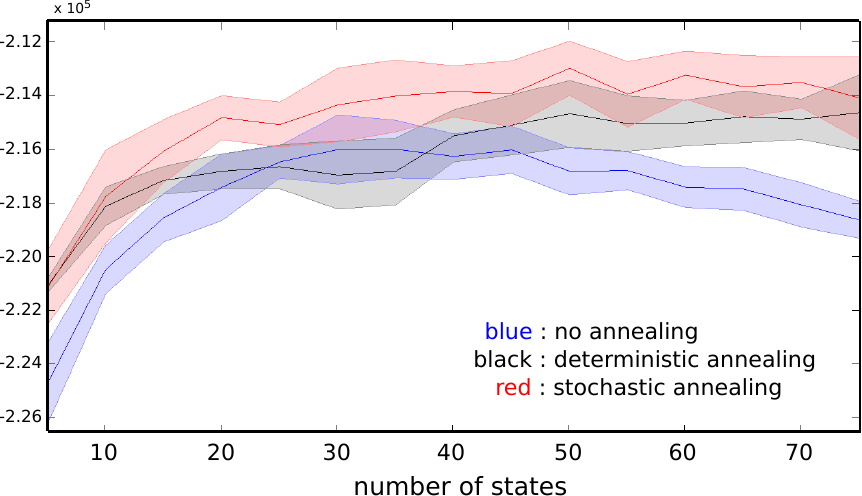}
 \caption{The variational objective function vs number of states for HMMs learned on quantized song sequences from the Million Song Dataset. We used 500 sequences of length 50 taken from 500 songs. In general, the models learned with annealing are closer to the true posterior than those without it. We also see that stochAVI performs better than detAVI on this problem.}\label{fig.msd_bound}
\end{figure}

\paragraph{Results.} In Figure \ref{fig.digit_bound} we show results of the variational lower bound for a 5-state and 10-state HMM learned from the characters dataset. As is evident, stochAVI consistently converges to a better posterior approximation than detAVI and VI. In Figure \ref{fig.msd_bound} we show the variational objective function for the MSD problem as a function of the number of states. Since we learn a joint HMM across songs, we find that a more complicated model with larger state space is better. Again we see that stochAVI outperforms detAVI, and that the annealing and non-annealing do not perfectly agree on the ideal number of states.

\subsection{The Gaussian mixture model}

\paragraph{Setup.} For the final experiment we evaluated the performance of stochAVI on an a $K$-state Gaussian mixture model (GMM). The parameters for this model are $\Theta = \{\pi,\mu_{1:K},\Lambda_{1:K}\}$, which includes the mixing weights $\pi$ and mean and precision for each Gaussian $(\mu_k,\Lambda_k)$. We select $q(\pi,\mu_{1:K},\Lambda_{1:K})=q(\pi)\prod_k q(\mu_k)q(\Lambda_k)$ as our factorization and set them to the same form as the prior, which is Dirichlet for $\pi$ and independent normal and Wishart distributions for $(\mu_k,\Lambda_k)$. We evaluate the three inference approaches on the MNIST digits dataset. For this problem, we first reduced the dimensionality by projecting the original $28 \times 28$ images onto their first 30 principal components. We then randomly selected 1,000 digits for each digit, 0 through 9, for training, and a separate 100 each for testing. We learned 50 different Gaussian mixture models for values of $K \in \{3,6,9,12,15,18\}$ for each digit, giving a total of 3,000 
experiments for each inference method.

\begin{figure}[t!]
\centering
 \subfigure[Gaussians mixture ($K=6$)]{\includegraphics[width=1\textwidth]{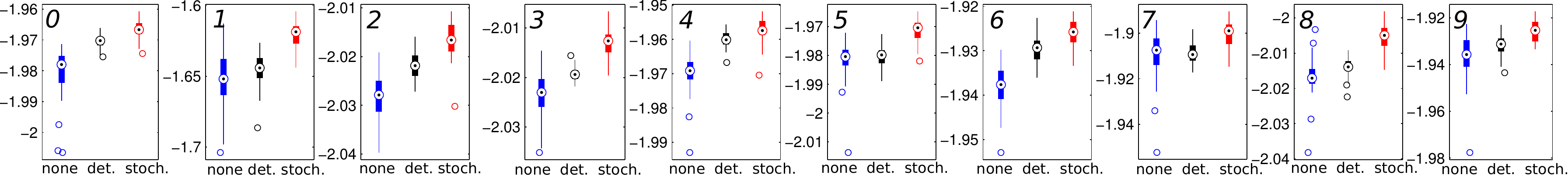}}
 \subfigure[Gaussians mixture ($K=12$)]{\includegraphics[width=1\textwidth]{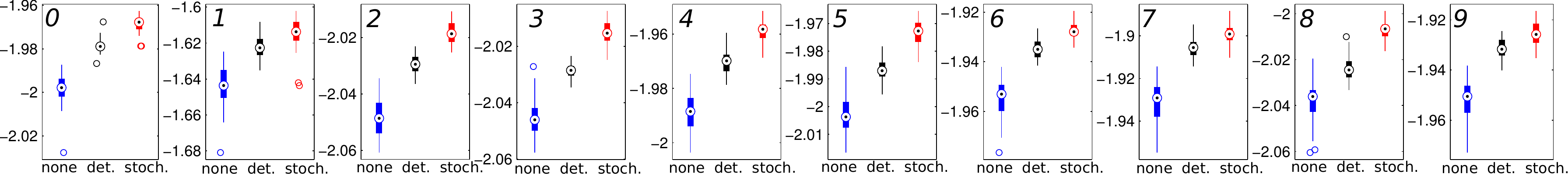}}
  \subfigure[Gaussians mixture ($K=18$)]{\includegraphics[width=1\textwidth]{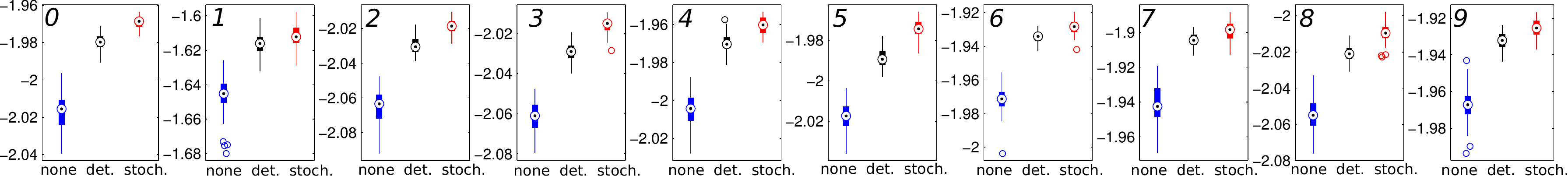}}
\caption{The variational objective function ($\times 10^{5}$) for each digit for a $6, 12$ and $18$ component Gaussian mixture model.}\label{fig.digitbox}
\end{figure}

\paragraph{Annealing.} Annealing for the GMM is more complicated in general than for LDA and the discrete HMM, which are restricted to Dirichlet-multinomial distributions. Annealing for $q(\pi)$ is straightforward, being a Dirichlet distribution, and follows the approach outlined above: We introduce a variational distribution on the hidden cluster assignments, where $\phi_n$ is a variational multinomial distribution on the cluster for observation $n$. Updating the variational parameter of $q(\pi)$ is then the same as LDA. The random vector $\eta_t$ at iteration $t$ with which this parameter is averaged corresponds to a random allocation of a dataset of the same size to the $K$ clusters.

We give a high level description for the more complicated $q(\mu_k)$ and $q(\Lambda_k)$ here. We use the allocation vector $\eta_t$ to scale the initializations for each Gaussian and then perform weighted averaging of the sufficient statistics from the data with those calculated from the initialization. Since we deterministically initialize each $q(\Lambda_k)$ to have an expectation of the empirical precision of the data, the update for $q(\Lambda_k)$ corresponds to taking the correct distribution on the precision $\Lambda_k$ and shrinking it towards the prior. In effect, after each iteration this stretches out the true update for the covariance of each Gaussian increasing it's ``reach,'' but in decreasing amounts as $\rho_t\rightarrow 0$. This is very similar to what is done by detAVI, with the exception that stochAVI incorporates randomness.

For $q(\mu_k)$ we randomly initialize the mean by drawing from a Gaussian with the empirical mean and covariance of the data. We initialize the precision to ten times the empirical precision of the data. Using this initialization in our annealing scheme corresponds to an update of $q(\mu_k)$ where the mean is approximately a linear combination of the empirical mean of the data assigned to cluster $k$ with a new randomly initialized mean. The covariance of $q(\mu_k)$ is approximately the true update to the covariance stretched out according to the prior. Both updates for $q(\mu_k)$ and $q(\Lambda_k)$ increase uncertainty, allowing these Gaussians to move around more in the initial iterations. We note that detAVI does not result in a modification to the mean of $q(\mu_k)$, and so this is an additional feature of stochastic annealing for the GMM.

\begin{figure}[t!]
\centering
 {\includegraphics[width=.6\textwidth]{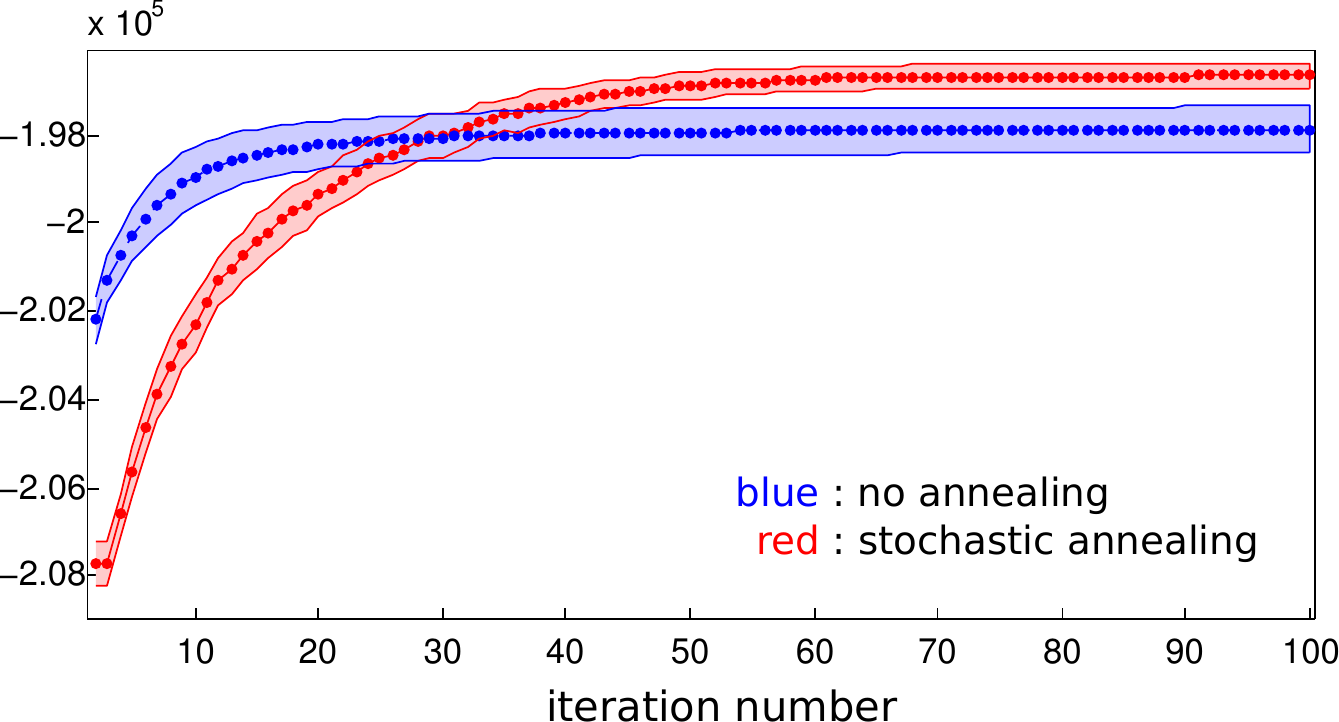}}
 \caption{Variational objective function vs iteration for 50 runs of a GMM with 6 components: (red) stochAVI, (blue) VI. We omit detAVI since it deforms the objective, and so is only comparable after annealing is turned off. Both stochAVI and VI are evaluated on the true variational objective function for each iteration. A similar pattern per iteration was observed with LDA and the HMM.}\label{fig.digitELBO}
\end{figure}
\paragraph{Results.} In Figure \ref{fig.digitbox} we show box plots of the variational objective as a function of number of Gaussians for the three methods. Again, stochAVI outperforms detAVI and VI in that it converges to a better local optimal solution. It also appears robust to model complexity in that the gap in performance grows with an increasing number of Gaussians.

In Table \ref{tab.digits} we show quantitative performance of a prediction task using the 100 testing examples for each digit. Using a naive Bayes classifier based on the mean of the learned $q$ distributions, we use average the classification accuracy for each value of $K$. Though the methods do not achieve state of the art performance on this dataset, the relative performance is more important here, where we see a slight improvement with stochAVI, followed by detAVI and then VI. This indicates that the increase improvement of $q$ can translate to an improvement in the end task, though the improvement is not major in this case.

In Figure \ref{fig.digitELBO} we plot the variational objective as a function of iteration for the digit 0. We see that stochAVI starts out with worse performance as it explores the space of the objective function, but then converges on a better local optimal solution. We omit detAVI since it deforms the variational objective function, meaning the curve actually decrease with iteration since the scale of the entropy is decreasing with each iteration.

\begin{table}
\caption{Bayes classification prediction accuracy averaged over digits 0 through 9. We observe a slight improvement in classification with effectively the same computation time.\vspace{2pt}}\label{tab.digits}\centering
\begin{threeparttable}
\begin{tabular}{ rccccc }
\toprule
Model & K=3 & K=6 & K=9 & K=12 & K=15  \\
\midrule
VI &  0.945  &  0.939  &  0.934 &   0.926 &   0.922\\
detAVI &  0.945  &  0.943  &  0.941  &  0.933  &  0.932\\
stochAVI &  0.947  &  0.944  &  0.943 &  0.938  &  0.936\\
\bottomrule
\end{tabular}
\end{threeparttable}
\end{table}

\section{Conclusion}
Variational inference is a valuable tool for scalable Bayesian inference, but convergence to a local optimal solution is a drawback that isn't satisfactorily addressed with multiple restarts. We have presented a method for variational inference based on simulated annealing that can help remove the need for these restarts by allowing for convergence to a better local optimal solution. The algorithm is based on a simple approach of averaging random initializations with parameter updates after each iteration in a way that favors randomness exploration at first, and gradually transitions to the correct, deterministic updates. We showed through empirical evaluation that annealing can have a benefit for several standard Bayesian models and compares favorably with existing deterministic annealing approaches.

\bibliographystyle{sjs}
\bibliography{mybib2}

\end{document}